\documentclass[10pt,letterpaper]{article}
\usepackage{arxiv}
\usepackage[T1]{fontenc}
\usepackage[utf8]{inputenc}
\usepackage{graphicx,booktabs,amsmath,amssymb}
\usepackage{adjustbox}
\usepackage[numbers,sort&compress]{natbib}
\usepackage{microtype}
\usepackage{placeins}
\usepackage[hidelinks]{hyperref}
\usepackage{xurl}
\usepackage{orcidlink}
\newcommand{\doi}[1]{\href{https://doi.org/#1}{\nolinkurl{#1}}}
\renewcommand{\shorttitle}{Multi-Model Fusion for Person Anomaly Search}
\renewcommand{\headeright}{A Preprint}
\renewcommand{\undertitle}{A Preprint}
\hypersetup{pdftitle={Synthetic-to-Real Text-Based Person Anomaly Search via Multi-Model Retrieval Fusion and Cross-Encoder Reranking},pdfauthor={Eugene Denteh and coauthors}}
\title{Synthetic-to-Real Text-Based Person Anomaly Search via Multi-Model Retrieval Fusion and Cross-Encoder Reranking}
\author{\normalfont
Eugene Denteh\textsuperscript{1,3}\orcidlink{0009-0001-7597-6732},
Blessing Agyei Kyem\textsuperscript{1,3}\orcidlink{0009-0006-6360-6386},
Andrews Danyo\textsuperscript{1,3}\orcidlink{0009-0008-3463-0349}\\[3pt]
Joshua Kofi Asamoah\textsuperscript{1,3}\orcidlink{0009-0002-3258-0479},
Richard Dzinyela\textsuperscript{2}\orcidlink{0000-0001-5114-5646},
Solomon Owusu-Ansah\textsuperscript{1,3}\\[3pt]
Dickson G. Asiedu\textsuperscript{4},
Armstrong Aboah\textsuperscript{1,3}\orcidlink{0000-0002-1605-1545}\\[9pt]
\normalfont\small\textsuperscript{1}North Dakota State University, Fargo, ND, USA\\
\normalfont\small\textsuperscript{2}Texas A\&M University, College Station, TX, USA\\
\normalfont\small\textsuperscript{3}Sustainable Mobility and Advanced Research in Transportation (SMART) Lab\\
\normalfont\small\textsuperscript{4}Ghana Institute of Management and Public Administration\\[5pt]
\normalfont\footnotesize\href{mailto:eugene.denteh@ndsu.edu}{eugene.denteh@ndsu.edu},
\href{mailto:armstrong.aboah@ndsu.edu}{armstrong.aboah@ndsu.edu}
}
\date{}
\begin{document}
\maketitle

\begin{abstract}
Text-based person anomaly search requires a model to retrieve individuals from a large image gallery using natural-language descriptions of both appearance and behavior. This task is challenging because model training relies on synthetic image-text pairs, whereas evaluation is performed on real surveillance images containing domain differences, near-duplicate frames, and many visually similar distractors. We present a multi-model retrieval and reranking framework designed to improve synthetic-to-real transfer. The framework first adapts Qwen3-VL-Embedding-8B to synthetic person descriptions using bidirectional contrastive learning, explicit negative sampling, false-negative masking, and domain randomization. It then combines the normalized scores of six complementary Qwen3-VL checkpoints with those of a frozen SigLIP2 model using four-view gallery encoding. The fused retrieval scores identify the strongest candidates, which are jointly evaluated with the query using Qwen3-VL-Reranker-8B. The cross-encoder and retrieval scores are subsequently blended to produce the final ranking. We also construct a distractor-matched validation protocol entirely from synthetic data to select checkpoint weights, fusion coefficients, candidate depths, and reranking parameters without accessing the official test images. The final system achieves \(94.7815\%\) mAP and \(90.8493\%\) R@1 under the official AI City Challenge 2026 Track~4 test setting, with a reported eighth-place finish on the public leaderboard.
\end{abstract}

\section{Introduction}

Text-based person anomaly search retrieves people from an image gallery using descriptions of both appearance and behavior~\cite{yang2025beyond,ju2025anomalylmm}. A useful match must satisfy clothing and physical-attribute constraints while also depicting the specified action, such as running, falling, or lying down. A visually similar person performing a different action is therefore a difficult distractor. The PAB benchmark makes this distinction explicit through synthetic training pairs and real-world evaluation images~\cite{yang2025beyond}. This setting raises a practical question: how can task-specific adaptation on synthetic data support reliable ranking in a real surveillance gallery?

Related transportation research provides several useful perspectives. Aboah's vision-based traffic anomaly detector~\cite{aboah2021anomaly} and DeepSegmenter~\cite{aboah2023deepsegmenter} study unusual events and driver actions, while QueryMotion combines attributes and motion reasoning for natural-language vehicle retrieval~\cite{asamoah2026querymotion}. These tasks emphasize the importance of behavioral evidence alongside appearance, although their datasets and evaluation objectives differ from person anomaly search. In the present task, domain differences, near-duplicate frames, and large numbers of plausible distractors create additional ranking challenges.

We address these challenges through a multi-stage framework. First, we adapt Qwen3-VL-Embedding-8B~\cite{qwen3vlembedding} to synthetic person descriptions using contrastive learning, explicit negative sampling, false-negative masking, and domain randomization. Second, we fuse standardized scores from six complementary Qwen3-VL checkpoints with a frozen SigLIP2 branch~\cite{tschannen2025siglip}. Third, Qwen3-VL-Reranker-8B jointly evaluates the query and the strongest image candidates, and its scores are blended with retrieval scores. A distractor-matched synthetic validation protocol selects the fusion and reranking settings.

The study makes three contributions:
\begin{enumerate}
\item A retrieval framework that combines task-adapted Qwen3-VL checkpoints, frozen SigLIP2 representations, and selective cross-encoder refinement for synthetic-to-real person anomaly search.
\item A synthetic validation protocol with a large gallery and identity-preserving, behavior-differing distractors, designed to expose errors obscured by nearly saturated conventional validation.
\item An evaluation of retrieval, training, fusion, and reranking configurations under the AI City Challenge 2026 Track~4 setting~\cite{Tang26AICity26}. The reported final result is $94.7815\%$ mAP and $90.8493\%$ R@1, placing SMART Lab eighth on the reported public leaderboard.
\end{enumerate}

The contribution is the task-specific integration and evaluation of these components. The results support the value of complementary retrieval evidence and focused candidate refinement; they do not establish that frozen and adapted models are inherently complementary in every setting. Here, ``synthetic-only training'' refers to task-specific adaptation, rather than the pretraining data of the foundation models.

\begin{figure}[tbp]
  \centering
  \includegraphics[width=0.70\linewidth]{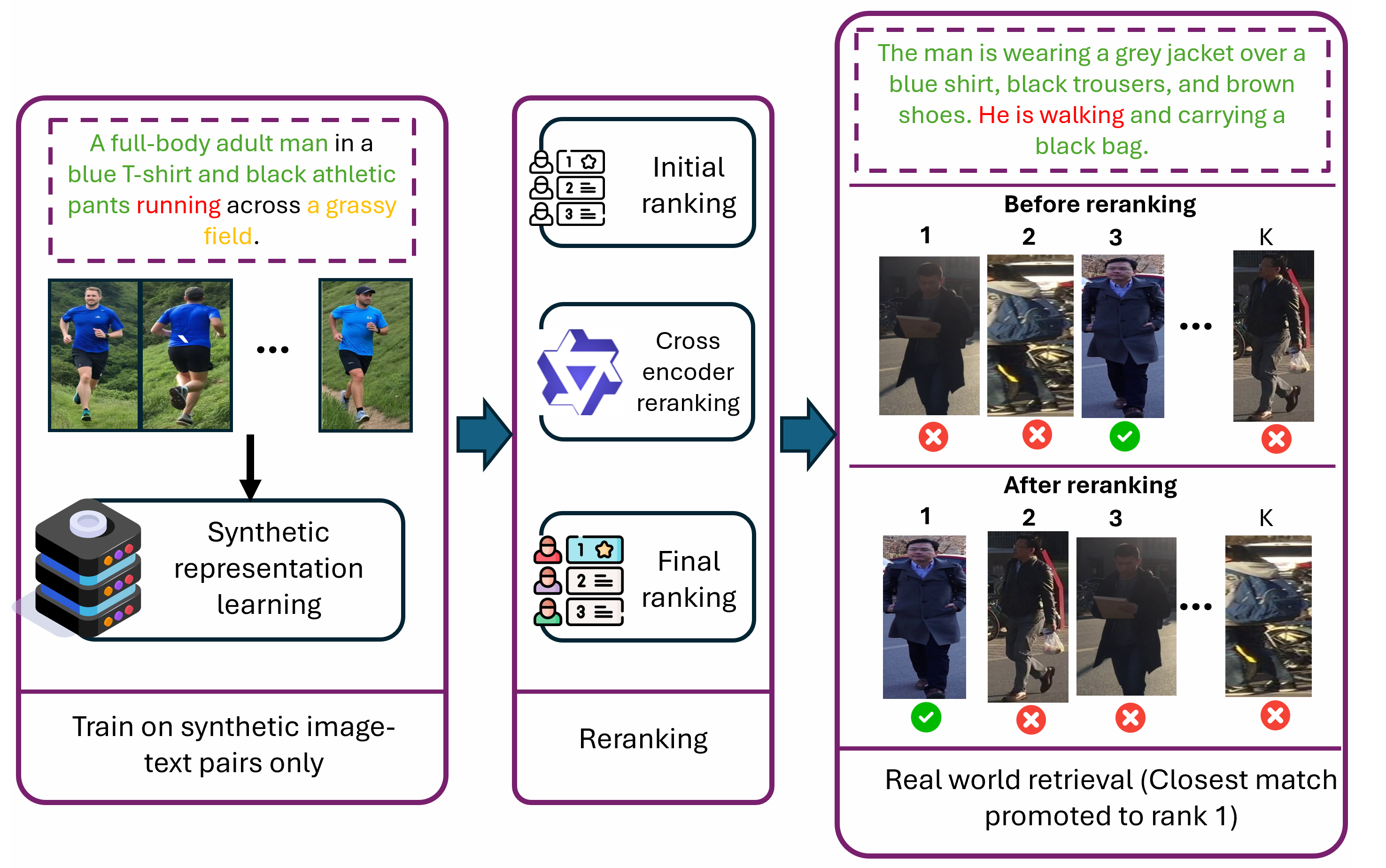}
  \caption{Synthetic-to-real text-based person anomaly retrieval setting. Given synthetic image-text pairs for training and a natural-language query at inference, the goal is to rank images from a real-world gallery according to their agreement with the described person and behavior. The examples show running and walking queries where multi-model retrieval fusion and cross-encoder reranking promote the best-matching real image to Rank~1 among visually similar distractors.}
  \label{fig:teaser}
\end{figure}

\section{Related Work}

\subsection{Vision-Language Pretraining}

CLIP~\cite{radford2021clip} learns transferable image-text representations through natural-language supervision, while ALIGN~\cite{jia2021align} scales dual-encoder contrastive learning using noisy image-text pairs. These models establish a shared-space retrieval paradigm in which gallery images can be encoded independently of a query. SigLIP~\cite{zhai2023siglip} replaces globally normalized contrastive training with a pairwise sigmoid objective, providing the foundation for the SigLIP2 branch used here. BLIP-2~\cite{li2023blip2} connects frozen image encoders and language models through a trainable querying transformer, illustrating a complementary approach to reusing pretrained components. These studies motivate separating general vision-language knowledge from task-specific adaptation, while leaving behavior-sensitive matching as an additional requirement.

\subsection{Text-Based Person Retrieval and Anomaly Search}

Li et al.~\cite{li2017person} introduced person search with natural-language descriptions and the CUHK-PEDES benchmark. SSAN~\cite{ding2021ssan} improves part-aware image-text alignment, while DSSL~\cite{zhu2021dssl} separates person information from surrounding visual content. IRRA~\cite{cvpr23crossmodal} learns implicit relations between visual and textual tokens. UFineBench~\cite{zuo2024ufinebench} extends evaluation to more detailed descriptions. Earlier cross-modal projection learning~\cite{zhang2018cmpm} uses matching and classification objectives to learn discriminative image-text embeddings. More recent approaches adapt the representation to subtle pedestrian details: CFine~\cite{yan2023cfine} transfers CLIP knowledge through fine-grained feature interaction, IVT~\cite{shu2022ivt} learns implicit alignment at sentence, phrase, and word levels, and PLIP~\cite{zuo2023plip} introduces person-specific pretraining tasks involving body regions, attributes, and identities. These methods highlight the need to preserve local semantic detail when transferring generic encoders to person search.

Training-pair reliability is another important consideration. RaSa~\cite{bai2023rasa} distinguishes strong and weak positive correspondences and learns sensitivity to replaced words. RDE~\cite{qin2024rde} addresses noisy image-text correspondences through consensus-based clean-pair selection and a robust alignment loss. Their treatment of imperfect correspondence is relevant to descriptions that match only some attributes of an image. These works provide appearance-sensitive retrieval foundations. Beyond Walking~\cite{yang2025beyond} introduces the PAB benchmark and a cross-modal pose-aware framework for retrieving both routine and anomalous behaviors. AnomalyLMM~\cite{ju2025anomalylmm} further studies behavioral saliency and knowledge-aware reranking on PAB. Our framework follows this behavior-sensitive task formulation and examines complementary embedding models and selective cross-encoder refinement.

\subsection{Behavior Understanding and Multimodal Transportation Models}

Behavior understanding has also been studied through temporal localization and multiple sensor views. DeepSegmenter~\cite{aboah2023deepsegmenter} combines segmentation and classification in untrimmed driving videos; SigSegment~\cite{aboah2023sigsegment} studies signal-based segmentation of anomalous driving behavior. Dontoh et al.~\cite{dontoh2025contextual} compare driver-facing and dual-view inputs for distraction detection. At the scene level, Denteh et al.~\cite{denteh2026congestion} combine flow-guided attention with empirical mode decomposition for congestion classification. Together, these studies motivate distinguishing behavioral evidence from static appearance; temporal or multi-view evidence is not evaluated by our image-gallery protocol.

Language-based interpretations offer a complementary direction. Shoman et al.~\cite{shoman2024captioning} study parallel dense video captioning for traffic-event analysis, and Kyem et al.~\cite{kyem2025dual} specialize separate vision-language models for traffic captioning and visual question answering. Muturi et al.~\cite{muturi2025spatial} use structured prompts for spatial relation reasoning, while QueryMotion~\cite{asamoah2026querymotion} combines attribute and motion cues for natural-language vehicle retrieval. These works support the relevance of explicit semantic and relational cues, but do not directly measure person anomaly retrieval on PAB.

\subsection{Synthetic Data, Visual Robustness, and Domain Adaptation}

Domain randomization~\cite{tobin2017domainrandomization} increases variability in simulated imagery to support real-world transfer. SimGAN~\cite{shrivastava2017simgan} instead refines synthetic images adversarially using unlabeled real images while preserving their annotations. For person re-identification, CamStyle~\cite{zhong2018camstyle} uses camera-style translation to augment training images and smooth cross-camera differences. These studies represent different responses to appearance shift; the synthetic validation protocol in this work also addresses the separate problem of selecting configurations before real-gallery evaluation.

Synthetic pedestrian imagery and domain adaptation were explored by Bak et al.~\cite{bak2018domain} and UnrealPerson~\cite{zhang2021unrealperson}. UniPT~\cite{shao2023unified} uses pseudo descriptions for task-specific person retrieval pretraining, and Cao et al.~\cite{cao2026empiricalstudyvalidatingsynthetic} systematically examine synthetic data for text-based person retrieval. MALS and its APTM framework~\cite{yang2023mals} further combine diffusion-generated person images, attribute prompts, and text matching, demonstrating a concrete route from synthetic multimodal pretraining to real person-retrieval benchmarks. In a related infrastructure setting, Kyem et al.~\cite{kyem2024weather} use weather-adaptive synthetic data for power-line inspection. These studies motivate examining how generated data and task-specific adaptation transfer to real imagery.

Robust traffic perception also depends on image conditions and sampling. Tran et al.~\cite{tran2024lowlight} combine low-light enhancement and ensemble learning for fisheye imagery; Owor et al.~\cite{owor2025fisheye} develop an ensemble detection pipeline for distorted surveillance images. Aboah et al.~\cite{aboah2023helmet} study few-shot data sampling for multi-class helmet-violation detection. Kyem et al.~\cite{kyem2025advancing} investigate locally collected pavement imagery, and PaveSync~\cite{kyem2025pavesync} consolidates heterogeneous pavement datasets. These are related examples of data diversity and visual robustness, rather than baselines for our retrieval task.

\subsection{Domain-Specific Adaptation, Fusion, and Candidate Refinement}

PaveCap~\cite{kyem2024pavecap} connects dense descriptions with pavement condition estimation, and PaveGPT~\cite{kyem2026pavegpt} studies domain-specific instruction tuning for infrastructure assessment. CrackGeoFM~\cite{kyem2026crack} combines a frozen visual backbone with task-specific adaptation for crack representation and topology preservation. Denteh et al.~\cite{denteh2025integrating,denteh2025demographics} integrate demographic, travel-behavior, and satellite-image information for urban forecasting. Although these applications differ from person retrieval, they provide broader examples of adapting or combining representations to satisfy domain-specific semantic and structural constraints.

Retaining pretrained knowledge during adaptation is also studied by WiSE-FT~\cite{wortsman2022wiseft}, which interpolates pretrained and fine-tuned model weights to improve robustness under distribution shift. Model soups~\cite{wortsman2022soups} average compatible fine-tuned weights to combine models without the inference overhead of a conventional ensemble. Both operate in parameter space, whereas our framework retains multiple encoders and combines their standardized similarity scores. This distinction is relevant to the accuracy--cost trade-off of the six-checkpoint retrieval ensemble.

Candidate refinement has a more direct retrieval precedent. Zhong et al.~\cite{zhong2017re} use reciprocal-neighborhood information for person re-identification, and SA-Person~\cite{xu2025sapersontextbasedpersonretrieval} incorporates scene context through a second-stage reranker. Qwen3-VL-Embedding and Qwen3-VL-Reranker~\cite{qwen3vlembedding} provide complementary embedding and cross-encoder models. Reciprocal rank fusion~\cite{cormack2009rrf} provides a rank-based way to combine retrieval lists without requiring comparable raw score scales. Our score standardization and weighted fusion use a different aggregation rule, followed by query-image cross-encoder refinement. The study evaluates how these model types can be combined with SigLIP2 and synthetic distractor-matched validation for behavior-sensitive person search.

\subsection{Negative Sampling and Model Selection under Domain Shift}

The quality of negative examples affects contrastive representation learning. Robinson et al.~\cite{robinson2021hardnegative} study controllable hard-negative sampling, while debiased contrastive learning~\cite{chuang2020debiased} corrects the contamination introduced when sampled negatives share the anchor's semantic label. Together, these studies explain why challenging negatives and false-negative handling should be considered jointly. In person anomaly search, the issue is particularly nuanced because the same identity can be relevant for one behavioral description and irrelevant for another; the masking and validation design must therefore respect the full query semantics.

DomainBed~\cite{gulrajani2021domainbed} emphasizes that model-selection criteria are an integral part of evaluating domain generalization algorithms. This observation motivates specifying the validation domain and candidate construction alongside the training procedure. Our distractor-matched synthetic validation is designed to expose ranking errors before real-gallery evaluation, and its empirical value is assessed separately from the representation and reranking components.

\section{Proposed Method}
\label{sec_method}

\subsection{Problem Formulation}
\label{sec_problem}

Given a natural-language query $q$ and an image gallery
$\mathcal{G}=\{I_i\}_{i=1}^{N}$, text-based person anomaly search retrieves
images that match both the person and the behavior described in the query,
including clothing, physical attributes, routine actions, or anomalous
activities like falling or lying down. The relevant images $\mathcal{G}_{q}^{+}
\subseteq\mathcal{G}$ must score higher than all distractors in
$\mathcal{G}\setminus\mathcal{G}_{q}^{+}$. The challenge lies in distinguishing
complete matches from partial ones: distractors may show the correct person
performing a different action, a different person with similar clothing, or
a near-duplicate frame from adjacent video moments. The model must therefore
jointly satisfy appearance and behavioral constraints, not just one or the other.

This objective is naturally formulated as embedding-space retrieval. An image
encoder $f_I(\cdot)$ and text encoder $f_T(\cdot)$ project gallery images and
the query into a shared space, producing normalized representations
$\mathbf{v}_i = f_I(I_i)/\lVert f_I(I_i)\rVert_2$ and
$\mathbf{u}_q = f_T(q)/\lVert f_T(q)\rVert_2$. Their correspondence is measured
using cosine similarity:
\begin{equation}
s(I_i,q)
=
\mathbf{v}_i^{\top}\mathbf{u}_q.
\label{eq_similarity}
\end{equation}

\begin{figure}[t]
    \centering
    \includegraphics[width=\textwidth]{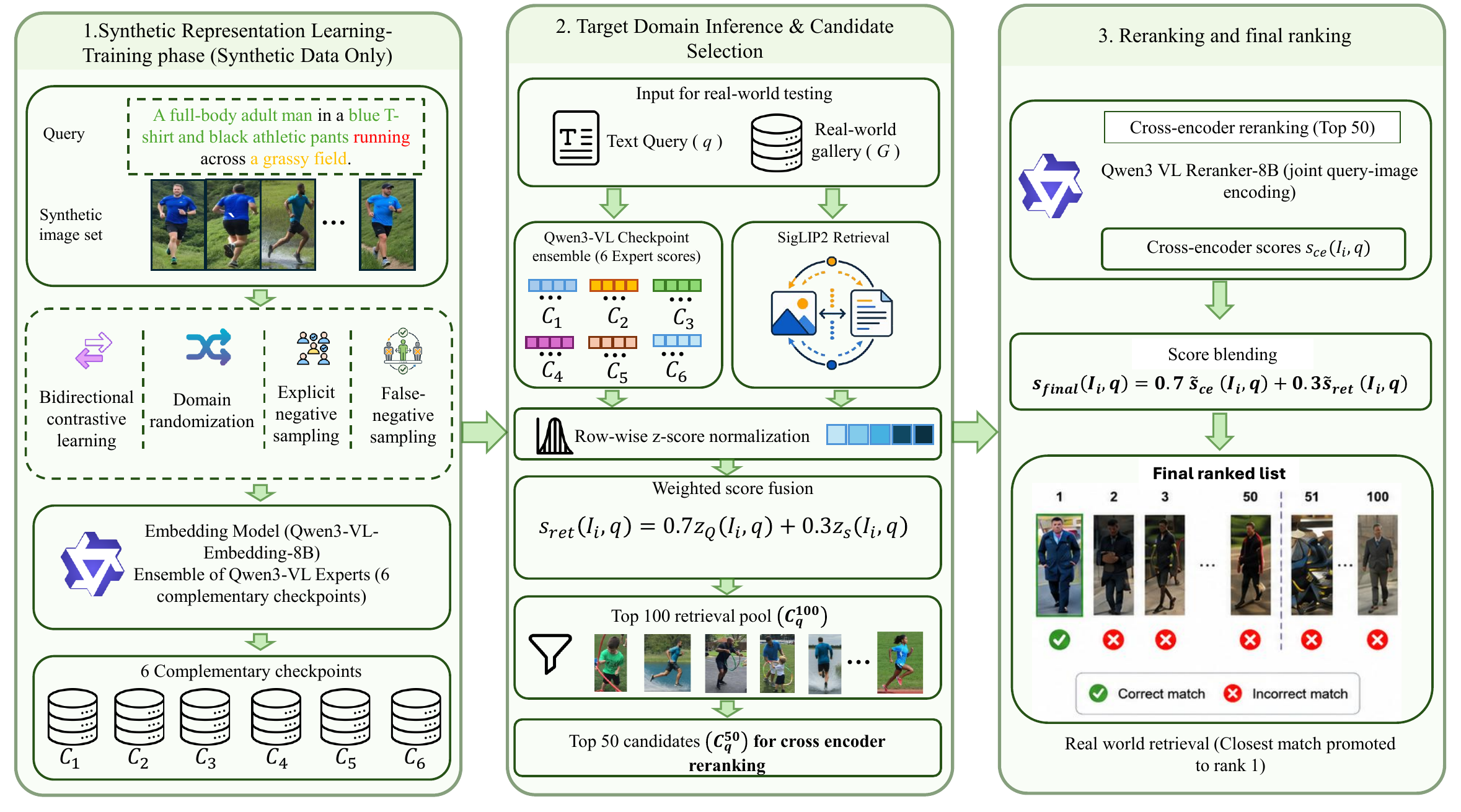}
    \caption{Overview of the proposed synthetic-to-real text-based person
    anomaly retrieval framework. Synthetic image-text pairs are used to
    adapt Qwen3-VL-Embedding-8B through bidirectional contrastive learning,
    explicit negative sampling, false-negative masking, and domain
    randomization, producing six complementary checkpoints. During
    inference, their weighted similarity scores are fused with scores from
    a frozen SigLIP2 retrieval branch using four-view gallery encoding. The
    combined retrieval score retains the top-100 candidates, of which the
    top-50 are jointly processed by Qwen3-VL-Reranker-8B and blended with
    the retrieval score to produce the final ranking. Distractor-matched
    synthetic validation is used for model and parameter selection without
    accessing the official challenge test images.}
    \label{fig:method_overview}
\end{figure}

The challenge in practice is that models are trained on synthetic images but
must retrieve from real surveillance galleries, creating a substantial domain
gap. Synthetic and real images differ in illumination, compression, background
complexity, image quality, and visual artifacts. Standard retrieval approaches
applied directly to this domain gap produce suboptimal results. We therefore
propose a three-stage framework: first, adapt Qwen3-VL-Embedding-8B to the
task using synthetic data; second, combine its checkpoints with a frozen
SigLIP2 branch to fuse complementary representations; and third, refine the
strongest candidates through joint query-image encoding via
Qwen3-VL-Reranker-8B. Figure~\ref{fig:method_overview} provides an overview
of this approach.

\subsection{Embedding Representation Learning}
\label{sec_embedding_learning}

The first stage learns a shared representation of person appearance and
behavior from synthetic PAB data~\cite{yang2025beyond}, consisting of
samples $(I_n,t_n,y_n,b_n)$ where $I_n$ is a synthetic image, $t_n$ is its
text description, and $y_n$ and $b_n$ denote the identity and behavior
labels.

Qwen3-VL-Embedding-8B serves as the embedding backbone~\cite{qwen3vlembedding}
and is adapted using LoRA~\cite{hu2022lora} with rank 16 and scaling factor
32, applied to the eligible linear layers. The visual encoder remains
frozen while the multimodal alignment layers remain trainable, preserving
the general visual representation while adapting the shared embedding space
to PAB descriptions.

Training uses a bidirectional contrastive objective with four explicit
negatives per sample. The text-to-image loss is
\begin{equation}
\mathcal{L}_{t2i}
=
-\frac{1}{B}
\sum_{n=1}^{B}
\log
\frac{
\exp\left(s(I_n^{+},t_n)/\tau\right)
}{
\exp\left(s(I_n^{+},t_n)/\tau\right)
+
\sum_{I\in\mathcal{N}_n}
m_{n,I}\exp\left(s(I,t_n)/\tau\right)
},
\label{eq_t2i_loss}
\end{equation}
where $B$ is the batch size, $\tau$ is the contrastive temperature, and
$m_{n,I}\in\{0,1\}$ masks candidates identified as possible false negatives
using a similarity margin of $0.02$. The image-to-text loss $\mathcal{L}_{i2t}$
follows the same formulation with the roles of image and text reversed, and
the complete embedding loss is their average,
$\mathcal{L}_{\mathrm{emb}} = \frac{1}{2}(\mathcal{L}_{t2i}+\mathcal{L}_{i2t})$.
This objective aligns matched image-text pairs while separating them from
candidates with different identities or behaviors.

Domain randomization varies illumination, contrast, color, blur, noise,
shadows, compression quality, crop scale, and horizontal orientation while
preserving identity and behavior labels, encouraging the model to learn
target semantics rather than synthetic visual patterns. Training and
continuation runs under different augmentation settings and random seeds
produce six complementary Qwen3-VL checkpoints. These checkpoints capture
different perspectives on the synthetic-to-real transfer and are fused in
the next stage via weighted ensemble.

\subsection{Multi-Model Retrieval Fusion}
\label{sec_fusion}

The second stage fuses the six Qwen3-VL checkpoints with a frozen SigLIP2~\cite{tschannen2025siglip}
model to combine specialized behavior-aware retrieval with general-purpose 
visual-language understanding. SigLIP2 is kept frozen to preserve its pretrained 
representations and maintain complementarity with the task-adapted Qwen3-VL 
ensemble. Adapting both models on the same synthetic data could reduce their diversity; the reported fusion experiments evaluate the benefit of retaining a frozen branch without establishing representation redundancy directly.

\subsubsection{Score Standardization:}

Since checkpoints and model families produce scores on different scales, we
standardize all scores to a common scale before fusion. For any score set
$\{a_i(q)\}_{i=1}^{N}$ from query $q$,
\begin{equation}
\mathcal{Z}_{q}\!\left(a_i(q)\right)
=
\frac{
a_i(q)-\mu_a(q)
}{
\sigma_a(q)+\epsilon
},
\label{eq_score_standardization}
\end{equation}
where $\mu_a(q)$ and $\sigma_a(q)$ are the gallery mean and standard deviation, and $\epsilon>0$ prevents division by zero.

\subsubsection{Qwen3-VL Checkpoint Ensemble:}

Each of the six Qwen3-VL checkpoints encodes both the original gallery image
and its horizontal reflection, which are averaged before scoring. The similarity
scores are standardized and combined via weighted fusion:
\begin{equation}
s_{Q}(I_i,q)
=
\sum_{m=1}^{M}
w_m \mathcal{Z}_q(s_m^{Q}(I_i,q)),
\qquad
w_m\geq0,
\quad
\sum_{m=1}^{M}w_m=1,
\label{eq_qwen_fusion}
\end{equation}
where $w_m$ is the validation-selected weight for checkpoint $m$ and $M=6$.

\subsubsection{Checkpoint Ensemble Strategy:}

The six checkpoints are trained on the same PAB dataset but under different
augmentation and random seeds, producing complementary representations that
capture different aspects of the synthetic-to-real transfer. Some checkpoints
may emphasize appearance details while others prioritize behavioral nuances.
Their complementarity reduces overfitting to spurious synthetic patterns and
provides more robust retrieval signals. The weighted fusion combines these
complementary perspectives, with weights $w_m$ selected via distractor-matched
validation to balance their contributions.

\subsubsection{SigLIP2 Four-View Encoding:}

The frozen SigLIP2 branch encodes four image views: the full image, its
horizontal flip, and upper and lower 60\% crops. These embeddings are averaged
and renormalized into a single gallery representation. The complete query is
processed by the SigLIP2 text encoder without paraphrasing, yielding a
similarity score $s_S(I_i,q)$ following Eq.~\ref{eq_similarity}.

\subsubsection{Cross-Model Score Fusion:}

The standardized Qwen3-VL ensemble score and the SigLIP2 score are combined as
\begin{equation}
s_{\mathrm{ret}}(I_i,q)
=
(1-\beta)\mathcal{Z}_q(s_Q(I_i,q))
+
\beta\mathcal{Z}_q(s_S(I_i,q)),
\label{eq_cross_model_fusion}
\end{equation}
where $\beta\in[0,1]$ controls the SigLIP2 contribution. Distractor-matched
validation selects $\beta=0.30$. The resulting retrieval score ranks the gallery
and provides candidates for the reranking stage.

\subsection{Candidate Reranking}
\label{sec_reranking}

The fused retrieval score efficiently ranks the gallery but may not fully
separate candidates with similar appearance, behavior, or scene context.
Detailed query-image comparison is therefore applied only to the strongest
candidates, reducing computational cost while focusing refinement where it
matters most.

For each query $q$, the top 100 retrieval-scored images form a candidate pool
$\mathcal{C}_q^{100}$. Of these, the top 50 images $\mathcal{C}_q^{50}$ are
processed by Qwen3-VL-Reranker-8B, while candidates ranked 51--100 retain their
retrieval order. The cross-encoder jointly encodes the query and each candidate
image, allowing direct visual-textual interaction that the dual encoders cannot
capture. This produces a compatibility score $s_{\mathrm{ce}}(I_i,q)$ reflecting
detailed agreement in appearance, action, body state, and scene context.

The retrieval and cross-encoder scores are independently standardized over
$\mathcal{C}_q^{50}$ and blended as
\begin{equation}
s_{\mathrm{final}}(I_i,q)
=
(1-\alpha)\widehat{s}_{\mathrm{ret}}(I_i,q)
+
\alpha\widehat{s}_{\mathrm{ce}}(I_i,q),
\qquad
I_i\in\mathcal{C}_q^{50},
\label{eq_final_blend}
\end{equation}
where $\alpha\in[0,1]$ controls the cross-encoder weight. Distractor-matched
validation selects $\alpha=0.70$. The top 50 candidates are reordered by
$s_{\mathrm{final}}$, while the remaining 51--100 retain their original retrieval
ranking.

\subsection{Distractor-Matched Validation}
\label{sec_validation}

Model and parameter selection require validation on data that reflects the
difficulty of the evaluation gallery. Conventional synthetic validation contains
fewer challenging distractors and produces nearly saturated scores, making model
selection less informative. We therefore propose a distractor-matched validation
protocol designed to surface ranking errors that occur on real data.

The protocol uses 1,978 held-out queries and 34,795 synthetic distractors,
forming a gallery of 36,773 images. Critically, 15\% of the distractors share
the query identity but depict different behaviors. These identity-preserving,
behavior-differing negatives test whether the model distinguishes actions rather
than relying solely on identity or clothing---a challenge that standard synthetic
validation underrepresents.

This protocol selects all model and pipeline parameters: the Qwen3-VL checkpoint
combination and weights $w_m$, the SigLIP2 fusion coefficient $\beta$, the
candidate pool depth (100) and reranking depth (50), and the cross-encoder blend
weight $\alpha$. Critically, the official challenge test images are never
accessed during training or validation, so model selection is reported to use synthetic validation only.

\subsection{Implementation Details}
\label{sec_implementation}

The Qwen3-VL embedding model is trained for 6,000 steps using AdamW with a
cosine schedule, a warm-up ratio of $0.05$, and weight decay of $0.1$. The
primary run uses a learning rate of $1\times10^{-7}$ and continuation runs
use $5\times10^{-8}$, with a contrastive temperature of $0.015$. Text
inputs are limited to 512 tokens and images use 32 to 64 visual tokens.

The SigLIP2 branch uses the pretrained
\texttt{google/siglip2-so400m-patch14-384} checkpoint without adaptation,
combined with the Qwen3-VL ensemble using $\beta=0.30$. The fused retrieval
model retains 100 candidates per query, of which the top 50 are sent to
Qwen3-VL-Reranker-8B, with a final blend of $\alpha=0.70$. All checkpoint
weights, fusion coefficients, candidate depths, and reranking parameters
remain fixed after validation.

\FloatBarrier

\section{Results and Discussion}
\label{sec_results}

We evaluate the proposed framework under the official AI City Challenge 2026 Track~4 setting \cite{Tang26AICity26}. The analysis first presents the overall challenge results and then examines the retrieval representations, training strategies, reranking methods, and distractor-matched validation protocol. All ablations use stage-matched comparisons so that the effect of each component can be interpreted independently.

\subsection{Official Challenge Results}
\label{sec_main_results}

Table~\ref{tab_main_results} compares the proposed framework with transferred configurations of the Cross-Modal Pose-aware framework, denoted CMP~\cite{yang2025beyond}. CMP configurations A and B are independent implementations, while the CMP consensus combines their rankings. The remaining configurations show the main stages of the proposed framework.

\begin{table}[t]
\centering\small
\caption{Performance under the official AI City Challenge 2026 Track~4 \cite{Tang26AICity26} test setting. All values are percentages.}
\label{tab_main_results}
\begin{adjustbox}{max width=\linewidth}
\begin{tabular}{lcccc}
\toprule
Method & mAP & R@1 & R@5 & R@10 \\
\midrule
CMP transfer configuration A
    & 67.2485 & 58.1901 & 78.3620 & 82.7098 \\
CMP transfer configuration B
    & 67.0983 & 57.9373 & 79.0698 & 82.0020 \\
CMP consensus ensemble
    & 85.4596 & 76.1881 & 97.5733 & 98.6350 \\
Qwen3-VL retrieval ensemble
    & 92.5510 & 87.3610 & 98.2305 & 98.9889 \\
Qwen3-VL ensemble + cross-encoder
    & 94.4586 & 90.3438 & 99.0394 & 99.4439 \\
Qwen3-VL + SigLIP2 fusion + cross-encoder
    & \textbf{94.7815} & \textbf{90.8493}
    & \textbf{99.0394} & \textbf{99.4944} \\
\bottomrule
\end{tabular}\end{adjustbox}
\end{table}

The CMP consensus substantially improves upon the two individual CMP configurations, showing that their rankings contain complementary information. However, the Qwen3-VL retrieval ensemble achieves a further increase to \(92.5510\%\) mAP and \(87.3610\%\) R@1. This comparison applies only to the transferred CMP configurations evaluated in this study and does not represent CMP performance under its original evaluation protocol.

Cross-encoder refinement increases mAP by \(1.9076\) points and R@1 by \(2.9828\) points over the Qwen3-VL retrieval ensemble. The larger improvement at Rank~1 indicates that joint query-image reasoning is particularly useful for reordering visually or behaviorally similar candidates near the top of the ranking.

Adding frozen SigLIP2 retrieval evidence produces the final result of \(94.7815\%\) mAP and \(90.8493\%\) R@1. This represents a further improvement of \(0.3229\) mAP points and \(0.5055\) R@1 points over Qwen3-VL reranking alone. The result shows that complementary retrieval evidence remains useful even after candidate-level refinement.

Table~\ref{tab_leaderboard} presents the reported public leaderboard snapshot. The SMART Lab submission is listed eighth. The organizer's published results summarize the leading five mAP entries~\cite{aicity2026results}.

\begin{table}[t]
\centering\small
\caption{Reported public leaderboard snapshot for AI City Challenge 2026 Track~4 \cite{Tang26AICity26}.}
\label{tab_leaderboard}
\begin{adjustbox}{max width=\linewidth}
\begin{tabular}{cclcccc}
\toprule
Rank & Team ID & Team Name & mAP & R@1 & R@5 & R@10 \\
\midrule
1  & 59  & Xiilab.AIpex
   & 99.3020 & 98.7361 & 99.9494 & 100.0000 \\
2  & 66  & Hallucination
   & 99.2551 & 98.7867 & 99.7472 & 99.7978 \\
3  & 9   & hiensumi
   & 98.3535 & 97.2194 & 99.6461 & 99.8989 \\
4  & 137 & UIT-OpenCubee
   & 95.9370 & 92.4166 & 99.8483 & 99.8483 \\
5  & 103 & Safe AI
   & 95.9250 & 93.7310 & 98.5844 & 99.5450 \\
6  & 53  & UWIPL
   & 95.3042 & 92.0121 & 98.8878 & 99.0394 \\
7  & 27  & VGU AI LAB
   & 95.1782 & 92.4671 & 98.4833 & 98.6855 \\
\textbf{8} & \textbf{97} & \textbf{SMART Lab}
   & \textbf{94.7815} & \textbf{90.8493}
   & \textbf{99.0394} & \textbf{99.4944} \\
9  & 174 & Cascade
   & 94.6485 & 90.7988 & 98.8878 & 99.3933 \\
10 & 127 & Team KODE
   & 94.5465 & 90.5460 & 98.9383 & 99.3933 \\
\bottomrule
\end{tabular}\end{adjustbox}
\end{table}

The final R@10 of \(99.4944\%\) shows that relevant images are usually retained within the first ten results. The main remaining challenge is therefore the precise ordering of the strongest candidates.

\subsection{Retrieval Representation and Fusion Analysis}
\label{sec_retrieval_fusion_results}

We next isolate the retrieval stage to examine the effects of task adaptation, checkpoint ensembling, validation-based weighting, and cross-model fusion. Table~\ref{tab_retrieval_fusion} reports all configurations before cross-encoder reranking.

\begin{table}[t]
\centering\small
\caption{Retrieval configurations evaluated before cross-encoder reranking. All values are percentages.}
\label{tab_retrieval_fusion}
\begin{adjustbox}{max width=\linewidth}
\begin{tabular}{lccc}
\toprule
Configuration & mAP & R@1 & R@10 \\
\midrule
SigLIP2, frozen standalone
    & 90.8875 & 84.0860 & 98.5990 \\
SigLIP2, task-adapted on PAB
    & 91.6301 & 85.5422 & 98.7991 \\
Individual Qwen3-VL checkpoint
    & 91.0915 & 86.2865 & 98.5990 \\
Qwen3-VL ensemble, unweighted
    & 92.0923 & 87.0872 & 98.8990 \\
Qwen3-VL ensemble, validation-weighted
    & 92.5510 & 87.3610 & 98.9889 \\
Qwen3-VL + frozen SigLIP2 fusion
    & \textbf{93.0933} & \textbf{88.0883} & \textbf{98.9991} \\
\bottomrule
\end{tabular}\end{adjustbox}
\end{table}

Task adaptation improves standalone SigLIP2 by \(0.7426\) mAP points and \(1.4562\) R@1 points, confirming that the synthetic PAB data provide useful task-specific information. However, both SigLIP2 configurations remain below the Qwen3-VL ensemble.

Combining six Qwen3-VL checkpoints improves mAP from \(91.0915\%\) to \(92.0923\%\). Validation-based weighting provides a further increase to \(92.5510\%\). These results show that checkpoint diversity and validation-guided weighting both contribute to the retrieval stage.

The strongest retrieval-only result is obtained by combining the weighted Qwen3-VL ensemble with frozen SigLIP2. Fusion improves mAP by \(0.5423\) points and R@1 by \(0.7273\) points over the weighted Qwen3-VL ensemble. Although frozen SigLIP2 is weaker as a standalone model, it contributes complementary retrieval evidence when combined with the task-adapted Qwen3-VL representation.

\subsection{Ablation Study}
\label{sec_ablation}

The following experiments examine embedding-stage training, reranked training, and candidate refinement. Each comparison keeps the remaining pipeline components fixed.

\subsubsection{Embedding-Stage Training Strategies}
\label{sec_embedding_training_ablation}

Table~\ref{tab_training_embedding} compares alternative training and augmentation strategies before reranking. The hard-negative fine-tuning configuration serves as the baseline.

\begin{table}[t]
\centering\small
\caption{Embedding-stage strategies evaluated before reranking. \(\Delta\)mAP is measured relative to the baseline.}
\label{tab_training_embedding}
\begin{adjustbox}{max width=\linewidth}
\begin{tabular}{lccc}
\toprule
Configuration & mAP & R@1 & $\Delta$ mAP \\
\midrule
Hard-negative fine-tuning baseline
    & 92.5510 & 87.3610 & --- \\
Query-paraphrase augmentation
    & 85.3432 & 77.8564 & -7.2078 \\
Query-paraphrase ensemble
    & 93.1906 & 88.8777 & +0.6396 \\
Iterative identity-based hard-negative mining
    & 89.1873 & 82.3054 & -3.3637 \\
Watermark-style augmentation
    & 92.2908 & 87.1587 & -0.2602 \\
\bottomrule
\end{tabular}\end{adjustbox}
\end{table}

Direct query-paraphrase augmentation produces the largest reduction, suggesting that some reformulations remove or alter behavior-related details. Combining the original and paraphrased queries avoids this loss and slightly exceeds the baseline, although the improvement remains modest.

Iterative identity-based hard-negative mining also reduces performance, indicating that excessive emphasis on synthetic identities can weaken transfer. Watermark-style augmentation has only a minor effect. The final embedding configuration therefore retains the original descriptions and a moderate hard-negative strategy.

\subsubsection{Reranked Training Strategies}
\label{sec_reranked_training_ablation}

Table~\ref{tab_training_reranked} compares training configurations after applying the same cross-encoder and score-blending procedure.

\begin{table}[t]
\centering\small
\caption{Training strategies evaluated with fixed reranking and score-blending settings.}
\label{tab_training_reranked}
\begin{adjustbox}{max width=\linewidth}
\begin{tabular}{lccc}
\toprule
Configuration & mAP & R@1 & $\Delta$ mAP \\
\midrule
Qwen3-VL ensemble + cross-encoder
    & 94.4586 & 90.3438 & --- \\
Domain randomization + compliant real-data mixture
    & 94.3297 & 90.2932 & -0.1289 \\
Identity-based hard-negative continuation
    & 94.2194 & 90.0910 & -0.2392 \\
\bottomrule
\end{tabular}\end{adjustbox}
\end{table}

Neither alternative improves upon the reranked Qwen3-VL baseline. The real-data mixture remains close to the baseline but provides no measurable gain, while identity-based continuation produces a larger reduction. These results indicate that more complex training mixtures or additional identity specialization do not necessarily improve real-gallery ranking.

\subsubsection{Candidate-Reranking Strategies}
\label{sec_reranking_ablation}

Table~\ref{tab_reranking_ablation} compares reranking methods using the same Qwen3-VL retrieval candidate pool.

\begin{table}[t]
\centering\small
\caption{Candidate-reranking strategies using a fixed Qwen3-VL retrieval pool.}
\label{tab_reranking_ablation}
\begin{adjustbox}{max width=\linewidth}
\begin{tabular}{lccc}
\toprule
Reranking configuration & mAP & R@1 & $\Delta$ mAP \\
\midrule
No reranking
    & 92.5510 & 87.3610 & --- \\
Zero-shot Qwen3-VL cross-encoder
    & 94.4586 & 90.3438 & +1.9076 \\
LoRA-adapted Qwen3-VL cross-encoder
    & 94.2240 & 90.0404 & +1.6730 \\
30B mixture-of-experts reranker
    & 94.1422 & 89.9899 & +1.5912 \\
Score blending, $\alpha\approx0.75$
    & \textbf{94.5316} & \textbf{90.5966} & \textbf{+1.9806} \\
\bottomrule
\end{tabular}\end{adjustbox}
\end{table}

The zero-shot cross-encoder improves mAP by \(1.9076\) points over the fixed retrieval baseline. In contrast, LoRA adaptation and the larger 30B reranker perform slightly worse. This result shows that greater model capacity or additional synthetic adaptation does not automatically improve candidate ordering.

Increasing the cross-encoder weight to approximately \(\alpha=0.75\) gives the strongest Qwen3-VL-only result. However, it remains \(0.2499\) mAP points below the complete Qwen3-VL--SigLIP2 system, confirming that the final gain cannot be attributed to reranker weighting alone.

\subsection{Effect of Distractor-Matched Validation}
\label{sec_validation_results}

The retrieval-fusion coefficient, cross-encoder blending weight, checkpoint weights, and candidate depths are selected using synthetic validation data without accessing the official challenge test images. Table~\ref{tab_validation_results} compares conventional and distractor-matched validation.

\begin{table}[t]
\centering\small
\caption{Diagnostic comparison of conventional and distractor-matched synthetic validation.}
\label{tab_validation_results}
\begin{adjustbox}{max width=\linewidth}
\begin{tabular}{lccccc}
\toprule
Protocol & Queries & Gallery & mAP & R@1 & R@10 \\
\midrule
Conventional validation
    & 14,722 & 14,722 & 98.6686 & 97.6973 & 99.9185 \\
Distractor-matched validation
    & 1,978 & 36,773 & 97.1995 & 95.3488 & 99.6461 \\
\bottomrule
\end{tabular}\end{adjustbox}
\end{table}

Conventional validation is nearly saturated. Increasing the gallery size and introducing more difficult distractors reduce mAP to \(97.1995\%\) and R@1 to \(95.3488\%\). The smaller reduction in R@10 shows that the added distractors primarily affect top-rank ordering rather than overall candidate retention.

The protocols use different query sets and gallery-construction procedures, so the numerical difference is interpreted only as a diagnostic comparison. The distractor-matched protocol provides a less saturated setting for model selection and more closely reproduces the large-gallery and high-distractor structure of the target task.

\subsection{Qualitative Analysis}
\label{sec_qualitative}

\begin{figure}[t]
    \centering
    \includegraphics[width=\textwidth]{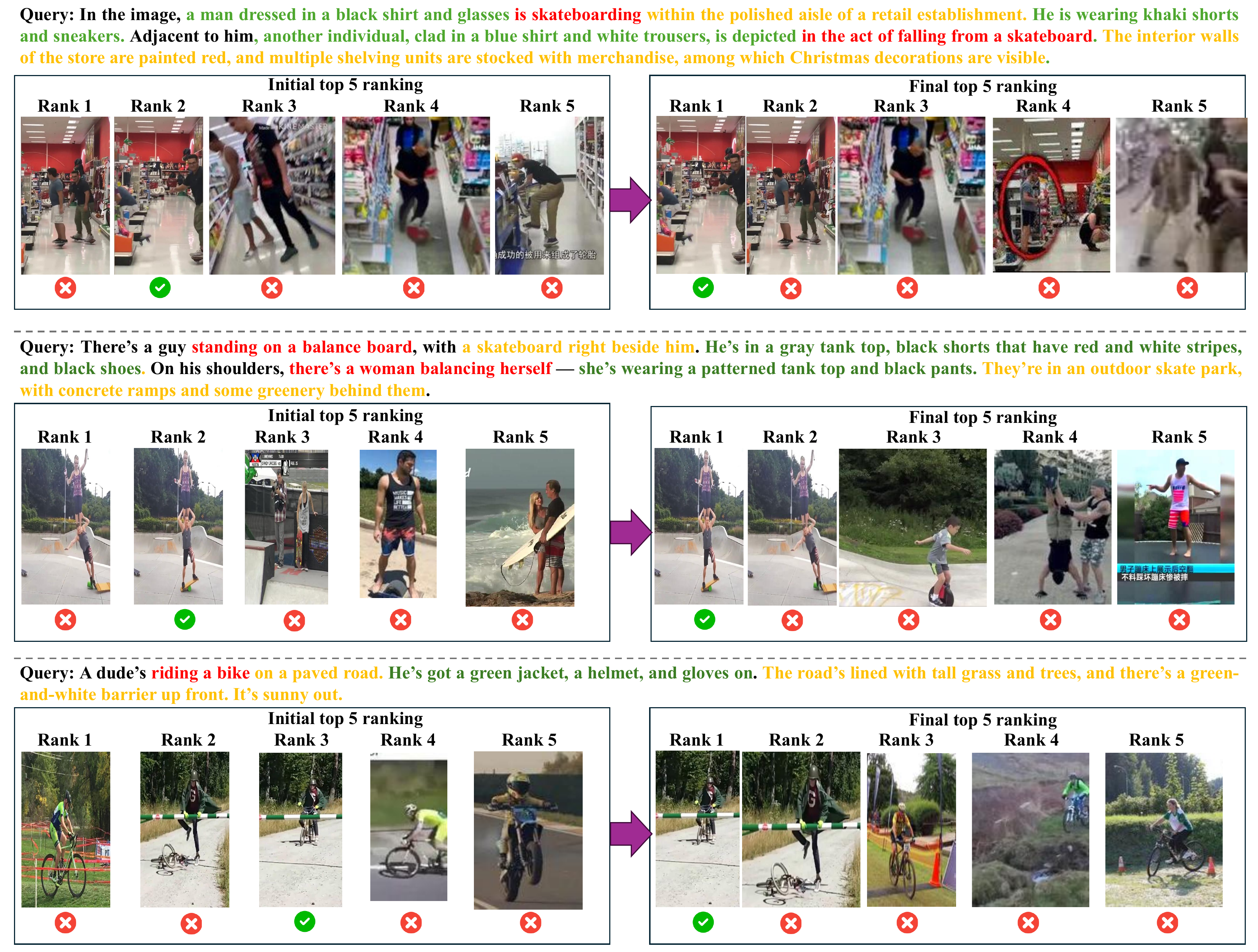}
    \caption{Qualitative retrieval results under the official AI City Challenge 2026 Track~4 setting \cite{Tang26AICity26}. Each row shows the query, the initial top-five ranking from Qwen3-VL--SigLIP2 fusion, and the final ranking after cross-encoder refinement. Green checkmarks indicate the annotated match, while red crosses indicate distractors.}
    \label{fig:qualitative_results}
\end{figure}

Figure~\ref{fig:qualitative_results} shows that the fused retrieval stage places the annotated image within the top three for all three examples. Cross-encoder refinement then promotes each match to Rank~1.

For Query~1, the description combines skateboarding with an interaction involving another person falling from a skateboard. The reranker promotes the candidate containing the complete behavioral context above images that mainly match the retail environment or individual appearance. Query~3 shows a similar improvement, as the final ranking gives greater weight to the cycling action together with the described clothing and scene.

Query~2 contains two nearly identical frames from the same skate-park sequence. The reranker selects the annotated frame, but the limited visual difference between the candidates illustrates the difficulty of exact frame-level retrieval. Overall, the examples show that the proposed task-specific combination of complementary retrieval representations and selective cross-encoder refinement improves appearance- and behavior-sensitive ordering while preserving a strong initial shortlist.

\subsection{Limitations}
The evaluation concerns one challenge setting, and the transferred CMP configurations are not a reproduction of CMP's original benchmark protocol. The reported comparisons do not include uncertainty estimates, so small differences should not be interpreted as statistically established improvements. The six-checkpoint ensemble and cross-encoder introduce computation and storage costs that are not quantified here. Exact checkpoint weights, batch size, hardware, random seeds, and split identifiers are not specified, limiting exact reproduction. The conventional and distractor-matched validation sets also differ in both queries and gallery construction. Finally, image-level retrieval does not explicitly model temporal evidence, and near-duplicate frames remain difficult. Task-specific synthetic adaptation does not imply that the foundation backbones were pretrained exclusively on synthetic data.

\FloatBarrier
\section{Conclusion}
\label{sec_conclusion}

We presented a multi-stage vision-language framework for synthetic-to-real text-based person anomaly search. The framework combines a Qwen3-VL retrieval ensemble with complementary SigLIP2 retrieval scores and applies cross-encoder reranking to the strongest gallery candidates. This design preserves efficient global retrieval while improving fine-grained agreement between the query and candidate images. The final system achieves \(94.7815\%\) mAP and \(90.8493\%\) R@1 under the official AI City Challenge 2026 Track~4 test setting, improving the Qwen3-VL retrieval baseline by \(2.2305\) and \(3.4883\) percentage points, respectively. The SMART Lab submission has a reported eighth-place finish on the public leaderboard. The results show that complementary retrieval representations and focused candidate reranking provide stronger transfer than increasing training complexity or reranker capacity alone. Future work will investigate temporal information from adjacent video frames and model distillation to improve near-duplicate discrimination while reducing inference cost.

\FloatBarrier
\bibliographystyle{unsrtnat}
\bibliography{references}

@InProceedings{yang2025beyond,
    author    = {Yang, Shuyu and Wang, Yaxiong and Zhu, Li and Zheng, Zhedong},
    title     = {{Beyond Walking: A Large-Scale Image-Text Benchmark for Text-based Person Anomaly Search}},
    booktitle = {Proceedings of the IEEE/CVF International Conference on Computer Vision (ICCV)},
    month     = {October},
    year      = {2025},
    pages     = {11720-11730}
,
note = {\href{https://openaccess.thecvf.com/content/ICCV2025/html/Yang_Beyond_Walking_A_Large-Scale_Image-Text_Benchmark_for_Text-based_Person_Anomaly_ICCV_2025_paper.html}{Open-access proceedings}}
}

@article{ju2025anomalylmm,
  title = {{AnomalyLMM: Bridging Generative Knowledge and Discriminative Retrieval for Text-Based Person Anomaly Search}},
  author = {Hao Ju and Hu Zhang and Zhedong Zheng},
  journal = {arXiv preprint arXiv:2509.04376},
  year = {2025},
  doi = {10.48550/arXiv.2509.04376}
}

@article{aboah2021anomaly,
  title = {{A Vision-based System for Traffic Anomaly Detection using Deep Learning and Decision Trees}},
  author = {Armstrong Aboah and Maged Shoman and Vishal Mandal and Sayedomidreza Davami and Yaw Adu-Gyamfi and Anuj Sharma},
  journal = {arXiv preprint arXiv:2104.06856},
  year = {2021},
  doi = {10.48550/arXiv.2104.06856}
}

@InProceedings{aboah2023deepsegmenter,
    author    = {Aboah, Armstrong and Bagci, Ulas and Mussah, Abdul Rashid and Owor, Neema Jakisa and Adu-Gyamfi, Yaw},
    title     = {{DeepSegmenter: Temporal Action Localization for Detecting Anomalies in Untrimmed Naturalistic Driving Videos}},
    booktitle = {Proceedings of the IEEE/CVF Conference on Computer Vision and Pattern Recognition (CVPR) Workshops},
    month     = {June},
    year      = {2023},
    pages     = {5359-5365}
,
note = {\href{https://openaccess.thecvf.com/content/CVPR2023W/AICity/html/Aboah_DeepSegmenter_Temporal_Action_Localization_for_Detecting_Anomalies_in_Untrimmed_Naturalistic_CVPRW_2023_paper.html}{Open-access proceedings}}
}

@article{asamoah2026querymotion,
  title = {{QueryMotion: A Modular Framework for Natural Language-Based Vehicle Retrieval via Attribute and Motion Reasoning}},
  author = {Joshua Kofi Asamoah and Blessing Agyei Kyem and Armstrong Aboah},
  journal = {IEEE Transactions on Intelligent Transportation Systems},
  volume = {27},
  number = {8},
  pages = {10069--10084},
  year = {2026},
  doi = {10.1109/TITS.2026.3680396}
}

@article{qwen3vlembedding,
  title = {{Qwen3-VL-Embedding and Qwen3-VL-Reranker: A Unified Framework for State-of-the-Art Multimodal Retrieval and Ranking}},
  author = {Mingxin Li and Yanzhao Zhang and Dingkun Long and Keqin Chen and Sibo Song and Shuai Bai and Zhibo Yang and Pengjun Xie and An Yang and Dayiheng Liu and Jingren Zhou and Junyang Lin},
  journal = {arXiv preprint arXiv:2601.04720},
  year = {2026},
  doi = {10.48550/arXiv.2601.04720}
}

@article{tschannen2025siglip,
  title = {{SigLIP 2: Multilingual Vision-Language Encoders with Improved Semantic Understanding, Localization, and Dense Features}},
  author = {Michael Tschannen and Alexey Gritsenko and Xiao Wang and Muhammad Ferjad Naeem and Ibrahim Alabdulmohsin and Nikhil Parthasarathy and Talfan Evans and Lucas Beyer and Ye Xia and Basil Mustafa and Olivier Hénaff and Jeremiah Harmsen and Andreas Steiner and Xiaohua Zhai},
  journal = {arXiv preprint arXiv:2502.14786},
  year = {2025},
  doi = {10.48550/arXiv.2502.14786}
}

@misc{Tang26AICity26,
  title = {{2026 Track 4: Text-Based Person Re-Identification (Sim2Real)}},
  author = {{AI City Challenge Organizers}},
  year = {2026},
  howpublished = {\url{https://www.aicitychallenge.org/2026-track4/}},
  note = {Accessed October 22, 2026}
}

@inproceedings{radford2021clip,
  title = {{Learning Transferable Visual Models From Natural Language Supervision}},
  author = {Alec Radford and Jong Wook Kim and Chris Hallacy and Aditya Ramesh and Gabriel Goh and Sandhini Agarwal and Girish Sastry and Amanda Askell and Pamela Mishkin and Jack Clark and Gretchen Krueger and Ilya Sutskever},
  booktitle = {International Conference on Machine Learning},
  year = {2021},
  pages = {8748--8763},
  note = {\href{https://proceedings.mlr.press/v139/radford21a.html}{Open-access proceedings}}
}

@inproceedings{jia2021align,
  title = {{Scaling Up Visual and Vision-Language Representation Learning With Noisy Text Supervision}},
  author = {Chao Jia and Yinfei Yang and Ye Xia and Yi-Ting Chen and Zarana Parekh and Hieu Pham and Quoc Le and Yun-Hsuan Sung and Zhen Li and Tom Duerig},
  booktitle = {International Conference on Machine Learning},
  year = {2021},
  pages = {4904--4916},
  note = {\href{https://proceedings.mlr.press/v139/jia21b.html}{Open-access proceedings}}
}

@inproceedings{zhai2023siglip,
  title = {{Sigmoid Loss for Language Image Pre-Training}},
  author = {Zhai, Xiaohua and Mustafa, Basil and Kolesnikov, Alexander and Beyer, Lucas},
  booktitle = {Proceedings of the IEEE/CVF International Conference on Computer Vision (ICCV)},
  year = {2023},
  pages = {11975--11986},
  note = {\href{https://openaccess.thecvf.com/content/ICCV2023/html/Zhai_Sigmoid_Loss_for_Language_Image_Pre-Training_ICCV_2023_paper.html}{Open-access proceedings}}
}

@inproceedings{li2023blip2,
  title = {{BLIP-2: Bootstrapping Language-Image Pre-training with Frozen Image Encoders and Large Language Models}},
  author = {Junnan Li and Dongxu Li and Silvio Savarese and Steven Hoi},
  booktitle = {International Conference on Machine Learning},
  year = {2023},
  pages = {19730--19742},
  note = {\href{https://proceedings.mlr.press/v202/li23q.html}{Open-access proceedings}}
}

@article{li2017person,
  title = {{Person Search with Natural Language Description}},
  author = {Shuang Li and Tong Xiao and Hongsheng Li and Bolei Zhou and Dayu Yue and Xiaogang Wang},
  journal = {arXiv preprint arXiv:1702.05729},
  year = {2017},
  doi = {10.48550/arXiv.1702.05729}
}

@article{ding2021ssan,
  title = {{Semantically Self-Aligned Network for Text-to-Image Part-aware Person Re-identification}},
  author = {Zefeng Ding and Changxing Ding and Zhiyin Shao and Dacheng Tao},
  journal = {arXiv preprint arXiv:2107.12666},
  year = {2021},
  doi = {10.48550/arXiv.2107.12666}
}

@article{zhu2021dssl,
  title = {{DSSL: Deep Surroundings-person Separation Learning for Text-based Person Retrieval}},
  author = {Aichun Zhu and Zijie Wang and Yifeng Li and Xili Wan and Jing Jin and Tian Wang and Fangqiang Hu and Gang Hua},
  journal = {arXiv preprint arXiv:2109.05534},
  year = {2021},
  doi = {10.48550/arXiv.2109.05534}
}

@InProceedings{cvpr23crossmodal,
    author    = {Jiang, Ding and Ye, Mang},
    title     = {{Cross-Modal Implicit Relation Reasoning and Aligning for Text-to-Image Person Retrieval}},
    booktitle = {Proceedings of the IEEE/CVF Conference on Computer Vision and Pattern Recognition (CVPR)},
    month     = {June},
    year      = {2023},
    pages     = {2787-2797}
,
note = {\href{https://openaccess.thecvf.com/content/CVPR2023/html/Jiang_Cross-Modal_Implicit_Relation_Reasoning_and_Aligning_for_Text-to-Image_Person_Retrieval_CVPR_2023_paper.html}{Open-access proceedings}}
}

@InProceedings{zuo2024ufinebench,
    author    = {Zuo, Jialong and Zhou, Hanyu and Nie, Ying and Zhang, Feng and Guo, Tianyu and Sang, Nong and Wang, Yunhe and Gao, Changxin},
    title     = {{UFineBench: Towards Text-based Person Retrieval with Ultra-fine Granularity}},
    booktitle = {Proceedings of the IEEE/CVF Conference on Computer Vision and Pattern Recognition (CVPR)},
    month     = {June},
    year      = {2024},
    pages     = {22010-22019}
}

@inproceedings{zhang2018cmpm,
  title = {{Deep Cross-Modal Projection Learning for Image-Text Matching}},
  author = {Zhang, Ying and Lu, Huchuan},
  booktitle = {Proceedings of the European Conference on Computer Vision (ECCV)},
  year = {2018},
  note = {\href{https://openaccess.thecvf.com/content_ECCV_2018/html/Ying_Zhang_Deep_Cross-Modal_Projection_ECCV_2018_paper.html}{Open-access proceedings}}
}

@article{yan2023cfine,
  title = {{CLIP-Driven Fine-Grained Text-Image Person Re-Identification}},
  author = {Yan, Shuanglin and Dong, Neng and Zhang, Liyan and Tang, Jinhui},
  journal = {IEEE Transactions on Image Processing},
  year = {2023},
  volume = {32},
  pages = {6032--6046},
  doi = {10.1109/TIP.2023.3327924}
}

@article{shu2022ivt,
  title = {{See Finer, See More: Implicit Modality Alignment for Text-based Person Retrieval}},
  author = {Xiujun Shu and Wei Wen and Haoqian Wu and Keyu Chen and Yiran Song and Ruizhi Qiao and Bo Ren and Xiao Wang},
  journal = {arXiv preprint arXiv:2208.08608},
  year = {2022},
  doi = {10.48550/arXiv.2208.08608}
}

@article{zuo2023plip,
  title = {{PLIP: Language-Image Pre-training for Person Representation Learning}},
  author = {Jialong Zuo and Jiahao Hong and Feng Zhang and Changqian Yu and Hanyu Zhou and Changxin Gao and Nong Sang and Jingdong Wang},
  journal = {arXiv preprint arXiv:2305.08386},
  year = {2023},
  doi = {10.48550/arXiv.2305.08386}
}

@inproceedings{bai2023rasa,
  title = {{RaSa: Relation and Sensitivity Aware Representation Learning for Text-based Person Search}},
  author = {Bai, Yang and Cao, Min and Gao, Daming and Cao, Ziqiang and Chen, Chen and Fan, Zhenfeng and Nie, Liqiang and Zhang, Min},
  booktitle = {Proceedings of the Thirty-Second International Joint Conference on Artificial Intelligence},
  year = {2023},
  pages = {555--563},
  doi = {10.24963/ijcai.2023/62}
}

@inproceedings{qin2024rde,
  title = {{Noisy-Correspondence Learning for Text-to-Image Person Re-identification}},
  author = {Qin, Yang and Chen, Yingke and Peng, Dezhong and Peng, Xi and Zhou, Joey Tianyi and Hu, Peng},
  booktitle = {Proceedings of the IEEE/CVF Conference on Computer Vision and Pattern Recognition (CVPR)},
  year = {2024},
  pages = {27197--27206},
  note = {\href{https://openaccess.thecvf.com/content/CVPR2024/html/Qin_Noisy-Correspondence_Learning_for_Text-to-Image_Person_Re-identification_CVPR_2024_paper.html}{Open-access proceedings}}
}

@article{aboah2023sigsegment,
  title = {{SigSegment: A Signal-Based Segmentation Algorithm for Identifying Anomalous Driving Behaviours in Naturalistic Driving Videos}},
  author = {Kelvin Kwakye and Younho Seong and Armstrong Aboah and Sun Yi},
  journal = {arXiv preprint arXiv:2304.09247},
  year = {2023},
  doi = {10.48550/arXiv.2304.09247}
}

@article{dontoh2025contextual,
  title = {{A Contextual Analysis of Driver-Facing and Dual-View Video Inputs for Distraction Detection in Naturalistic Driving Environments}},
  author = {Anthony Dontoh and Stephanie Ivey and Armstrong Aboah},
  journal = {arXiv preprint arXiv:2512.20025},
  year = {2025},
  doi = {10.48550/arXiv.2512.20025}
}

@article{denteh2026congestion,
  title = {{Hybrid Congestion Classification Framework Using Flow-Guided Attention and Empirical Mode Decomposition}},
  author = {Eugene Kofi Okrah Denteh and Blessing Agyei Kyem and Joshua Kofi Asamoah and Armstrong Aboah},
  journal = {arXiv preprint arXiv:2605.04752},
  year = {2026},
  doi = {10.48550/arXiv.2605.04752}
}

@article{shoman2024captioning,
  title = {{Enhancing Traffic Safety with Parallel Dense Video Captioning for End-to-End Event Analysis}},
  author = {Maged Shoman and Dongdong Wang and Armstrong Aboah and Mohamed Abdel-Aty},
  journal = {arXiv preprint arXiv:2404.08229},
  year = {2024},
  doi = {10.48550/arXiv.2404.08229}
}

@InProceedings{kyem2025dual,
    author    = {Kyem, Blessing Agyei and Owor, Neema J. and Danyo, Andrews and Asamoah, Joshua K. and Denteh, Eugene and Muturi, Tanner and Dontoh, Anthony and Adu-Gyamfi, Yaw and Aboah, Armstrong},
    title     = {{Task-Specific Dual-Model Framework for Comprehensive Traffic Safety Video Description and Analysis}},
    booktitle = {Proceedings of the IEEE/CVF International Conference on Computer Vision (ICCV) Workshops},
    month     = {October},
    year      = {2025},
    pages     = {5384-5392}
,
note = {\href{https://openaccess.thecvf.com/content/ICCV2025W/AICity/html/Kyem_Task-Specific_Dual-Model_Framework_for_Comprehensive_Traffic_Safety_Video_Description_and_ICCVW_2025_paper.html}{Open-access proceedings}}
}

@InProceedings{muturi2025spatial,
    author    = {Muturi, Tanner W. and Kyem, Blessing Agyei and Asamoah, Joshua Kofi and Owor, Neema Jakisa and Dzinyela, Richard and Danyo, Andrews and Adu-Gyamfi, Yaw and Aboah, Armstrong},
    title     = {{Prompt-Guided Spatial Understanding with RGB-D Transformers for Fine-Grained Object Relation Reasoning}},
    booktitle = {Proceedings of the IEEE/CVF International Conference on Computer Vision (ICCV) Workshops},
    month     = {October},
    year      = {2025},
    pages     = {5339-5347}
,
note = {\href{https://openaccess.thecvf.com/content/ICCV2025W/AICity/html/Muturi_Prompt-Guided_Spatial_Understanding_with_RGB-D_Transformers_for_Fine-Grained_Object_Relation_ICCVW_2025_paper.html}{Open-access proceedings}}
}

@article{tobin2017domainrandomization,
  title = {{Domain Randomization for Transferring Deep Neural Networks from Simulation to the Real World}},
  author = {Josh Tobin and Rachel Fong and Alex Ray and Jonas Schneider and Wojciech Zaremba and Pieter Abbeel},
  journal = {arXiv preprint arXiv:1703.06907},
  year = {2017},
  doi = {10.48550/arXiv.1703.06907}
}

@inproceedings{shrivastava2017simgan,
  title = {{Learning From Simulated and Unsupervised Images Through Adversarial Training}},
  author = {Shrivastava, Ashish and Pfister, Tomas and Tuzel, Oncel and Susskind, Joshua and Wang, Wenda and Webb, Russell},
  booktitle = {Proceedings of the IEEE Conference on Computer Vision and Pattern Recognition (CVPR)},
  year = {2017},
  note = {\href{https://openaccess.thecvf.com/content_cvpr_2017/html/Shrivastava_Learning_From_Simulated_CVPR_2017_paper.html}{Open-access proceedings}}
}

@inproceedings{zhong2018camstyle,
  title = {{Camera Style Adaptation for Person Re-Identification}},
  author = {Zhong, Zhun and Zheng, Liang and Zheng, Zhedong and Li, Shaozi and Yang, Yi},
  booktitle = {Proceedings of the IEEE Conference on Computer Vision and Pattern Recognition (CVPR)},
  year = {2018},
  note = {\href{https://openaccess.thecvf.com/content_cvpr_2018/html/Zhong_Camera_Style_Adaptation_CVPR_2018_paper.html}{Open-access proceedings}}
}

@InProceedings{bak2018domain,
author = {Bak, Slawomir and Carr, Peter and Lalonde, Jean-Francois},
title = {{Domain Adaptation through Synthesis for Unsupervised Person Re-identification}},
booktitle = {Proceedings of the European Conference on Computer Vision (ECCV)},
month = {September},
year = {2018}
,
note = {\href{https://openaccess.thecvf.com/content_ECCV_2018/html/Slawomir_Bak_Domain_Adaptation_through_ECCV_2018_paper.html}{Open-access proceedings}}
}

@InProceedings{zhang2021unrealperson,
    author    = {Zhang, Tianyu and Xie, Lingxi and Wei, Longhui and Zhuang, Zijie and Zhang, Yongfei and Li, Bo and Tian, Qi},
    title     = {{UnrealPerson: An Adaptive Pipeline Towards Costless Person Re-Identification}},
    booktitle = {Proceedings of the IEEE/CVF Conference on Computer Vision and Pattern Recognition (CVPR)},
    month     = {June},
    year      = {2021},
    pages     = {11506-11515}
,
note = {\href{https://openaccess.thecvf.com/content/CVPR2021/html/Zhang_UnrealPerson_An_Adaptive_Pipeline_Towards_Costless_Person_Re-Identification_CVPR_2021_paper.html}{Open-access proceedings}}
}

@InProceedings{shao2023unified,
    author    = {Shao, Zhiyin and Zhang, Xinyu and Ding, Changxing and Wang, Jian and Wang, Jingdong},
    title     = {{Unified Pre-Training with Pseudo Texts for Text-To-Image Person Re-Identification}},
    booktitle = {Proceedings of the IEEE/CVF International Conference on Computer Vision (ICCV)},
    month     = {October},
    year      = {2023},
    pages     = {11174-11184}
,
note = {\href{https://openaccess.thecvf.com/content/ICCV2023/html/Shao_Unified_Pre-Training_with_Pseudo_Texts_for_Text-To-Image_Person_Re-Identification_ICCV_2023_paper.html}{Open-access proceedings}}
}

@article{cao2026empiricalstudyvalidatingsynthetic,
  title = {{An Empirical Study of Validating Synthetic Data for Text-Based Person Retrieval}},
  author = {Min Cao and Yuxin Lu and Ziyin Zeng and Dong Yi and Jinqiao Wang and Mang Ye},
  journal = {arXiv preprint arXiv:2503.22171},
  year = {2025},
  doi = {10.48550/arXiv.2503.22171}
}

@inproceedings{yang2023mals,
  title = {{Towards Unified Text-based Person Retrieval: A Large-scale Multi-Attribute and Language Search Benchmark}},
  author = {Yang, Shuyu and Zhou, Yinan and Zheng, Zhedong and Wang, Yaxiong and Zhu, Li and Wu, Yujiao},
  booktitle = {Proceedings of the 31st ACM International Conference on Multimedia},
  year = {2023},
  pages = {4492--4501},
  doi = {10.1145/3581783.3611709}
}

@article{kyem2024weather,
  title = {{Weather-Adaptive Synthetic Data Generation for Enhanced Power Line Inspection Using StarGAN}},
  author = {Blessing Agyei Kyem and Joshua Kofi Asamoah and Ying Huang and Armstrong Aboah},
  journal = {IEEE Access},
  volume = {12},
  pages = {193882--193901},
  year = {2024},
  doi = {10.1109/ACCESS.2024.3520120}
}

@InProceedings{tran2024lowlight,
    author    = {Tran, Dai Quoc and Aboah, Armstrong and Jeon, Yuntae and Shoman, Maged and Park, Minsoo and Park, Seunghee},
    title     = {{Low-Light Image Enhancement Framework for Improved Object Detection in Fisheye Lens Datasets}},
    booktitle = {Proceedings of the IEEE/CVF Conference on Computer Vision and Pattern Recognition (CVPR) Workshops},
    month     = {June},
    year      = {2024},
    pages     = {7056-7065}
,
note = {\href{https://openaccess.thecvf.com/content/CVPR2024W/AICity/html/Tran_Low-Light_Image_Enhancement_Framework_for_Improved_Object_Detection_in_Fisheye_CVPRW_2024_paper.html}{Open-access proceedings}}
}

@InProceedings{owor2025fisheye,
    author    = {Owor, Neema Jakisa and Asamoah, Joshua Kofi and Muturi, Tanner Wambui and Owor, Anneliese Jakisa and Kyem, Blessing Agyei and Danyo, Andrews and Adu-Gyamfi, Yaw and Aboah, Armstrong},
    title     = {{A Unified Detection Pipeline for Robust Object Detection in Fisheye-Based Traffic Surveillance}},
    booktitle = {Proceedings of the IEEE/CVF International Conference on Computer Vision (ICCV) Workshops},
    month     = {October},
    year      = {2025},
    pages     = {5314-5321}
,
note = {\href{https://openaccess.thecvf.com/content/ICCV2025W/AICity/html/Owor_A_Unified_Detection_Pipeline_for_Robust_Object_Detection_in_Fisheye-Based_ICCVW_2025_paper.html}{Open-access proceedings}}
}

@InProceedings{aboah2023helmet,
    author    = {Aboah, Armstrong and Wang, Bin and Bagci, Ulas and Adu-Gyamfi, Yaw},
    title     = {{Real-Time Multi-Class Helmet Violation Detection Using Few-Shot Data Sampling Technique and YOLOv8}},
    booktitle = {Proceedings of the IEEE/CVF Conference on Computer Vision and Pattern Recognition (CVPR) Workshops},
    month     = {June},
    year      = {2023},
    pages     = {5350-5358}
,
note = {\href{https://openaccess.thecvf.com/content/CVPR2023W/AICity/html/Aboah_Real-Time_Multi-Class_Helmet_Violation_Detection_Using_Few-Shot_Data_Sampling_Technique_CVPRW_2023_paper.html}{Open-access proceedings}}
}

@article{kyem2025advancing,
  title = {{Advancing Pavement Distress Detection in Developing Countries: A Novel Deep Learning Approach with Locally-Collected Datasets}},
  author = {Blessing Agyei Kyem and Eugene Kofi Okrah Denteh and Joshua Kofi Asamoah and Kenneth Adomako Tutu and Armstrong Aboah},
  journal = {arXiv preprint arXiv:2408.05649},
  year = {2024},
  doi = {10.48550/arXiv.2408.05649}
}

@article{kyem2025pavesync,
  title = {{PaveSync: A Unified and Comprehensive Dataset for Pavement Distress Analysis and Classification}},
  author = {Blessing Agyei Kyem and Joshua Kofi Asamoah and Anthony Dontoh and Andrews Danyo and Eugene Denteh and Armstrong Aboah},
  journal = {arXiv preprint arXiv:2512.20011},
  year = {2025},
  doi = {10.48550/arXiv.2512.20011}
}

@article{kyem2024pavecap,
  title = {{PaveCap: A Multimodal Framework for Comprehensive Pavement Condition Assessment with Dense Captioning and PCI Estimation}},
  author = {Blessing Agyei Kyem and Eugene Denteh and Joshua Kofi Asamoah and Denver Tolliver and Armstrong Aboah},
  journal = {Journal of Transportation Engineering, Part B: Pavements},
  volume = {152},
  number = {2},
  year = {2026},
  doi = {10.1061/JPEODX.PVENG-1945}
}

@article{kyem2026pavegpt,
  title = {{Vision-Language Foundation Models for Comprehensive Automated Pavement Condition Assessment}},
  author = {Blessing Agyei Kyem and Joshua Kofi Asamoah and Anthony Dontoh and Armstrong Aboah},
  journal = {arXiv preprint arXiv:2604.08212},
  year = {2026},
  doi = {10.48550/arXiv.2604.08212}
}

@article{kyem2026crack,
  title = {{Multi-Task Crack Foundation Model for Engineering-Reliable Crack Representation and Topology Preservation in Civil Infrastructure}},
  author = {Blessing Agyei Kyem and Joshua Kofi Asamoah and Eugene Denteh and Armstrong Aboah},
  journal = {arXiv preprint arXiv:2606.05641},
  year = {2026},
  doi = {10.48550/arXiv.2606.05641}
}

@article{denteh2025integrating,
  title = {{Integrating Travel Behavior Forecasting and Generative Modeling for Predicting Future Urban Mobility and Spatial Transformations}},
  author = {Eugene Denteh and Andrews Danyo and Joshua Kofi Asamoah and Blessing Agyei Kyem and Twitchell Addai and Armstrong Aboah},
  journal = {arXiv preprint arXiv:2503.21158},
  year = {2025},
  doi = {10.48550/arXiv.2503.21158}
}

@article{denteh2025demographics,
  title = {{Demographics-Informed Neural Network for Multi-Modal Spatiotemporal forecasting of Urban Growth and Travel Patterns Using Satellite Imagery}},
  author = {Eugene Kofi Okrah Denteh and Andrews Danyo and Joshua Kofi Asamoah and Blessing Agyei Kyem and Armstrong Aboah},
  journal = {arXiv preprint arXiv:2506.12456},
  year = {2025},
  doi = {10.48550/arXiv.2506.12456}
}

@inproceedings{wortsman2022wiseft,
  title = {{Robust Fine-Tuning of Zero-Shot Models}},
  author = {Wortsman, Mitchell and Ilharco, Gabriel and Kim, Jong Wook and Li, Mike and Kornblith, Simon and Roelofs, Rebecca and Lopes, Raphael Gontijo and Hajishirzi, Hannaneh and Farhadi, Ali and Namkoong, Hongseok and Schmidt, Ludwig},
  booktitle = {Proceedings of the IEEE/CVF Conference on Computer Vision and Pattern Recognition (CVPR)},
  year = {2022},
  pages = {7959--7971},
  note = {\href{https://openaccess.thecvf.com/content/CVPR2022/html/Wortsman_Robust_Fine-Tuning_of_Zero-Shot_Models_CVPR_2022_paper.html}{Open-access proceedings}}
}

@inproceedings{wortsman2022soups,
  title = {{Model soups: averaging weights of multiple fine-tuned models improves accuracy without increasing inference time}},
  author = {Mitchell Wortsman and Gabriel Ilharco and Samir Ya Gadre and Rebecca Roelofs and Raphael Gontijo-Lopes and Ari S Morcos and Hongseok Namkoong and Ali Farhadi and Yair Carmon and Simon Kornblith and Ludwig Schmidt},
  booktitle = {International Conference on Machine Learning},
  year = {2022},
  pages = {23965--23998},
  note = {\href{https://proceedings.mlr.press/v162/wortsman22a.html}{Open-access proceedings}}
}

@InProceedings{zhong2017re,
author = {Zhong, Zhun and Zheng, Liang and Cao, Donglin and Li, Shaozi},
title = {{Re-Ranking Person Re-Identification With k-Reciprocal Encoding}},
booktitle = {Proceedings of the IEEE Conference on Computer Vision and Pattern Recognition (CVPR)},
month = {July},
year = {2017}
}

@article{xu2025sapersontextbasedpersonretrieval,
  title = {{SA-Person: Text-Based Person Retrieval with Scene-aware Re-ranking}},
  author = {Yingjia Xu and Jinlin Wu and Daming Gao and Zhen Chen and Yang Yang and Min Cao and Mang Ye and Zhen Lei},
  journal = {arXiv preprint arXiv:2505.24466},
  year = {2025},
  doi = {10.48550/arXiv.2505.24466}
}

@inproceedings{cormack2009rrf,
  title = {{Reciprocal rank fusion outperforms condorcet and individual rank learning methods}},
  author = {Cormack, Gordon V. and Clarke, Charles L A and Buettcher, Stefan},
  booktitle = {Proceedings of the 32nd international ACM SIGIR conference on Research and development in information retrieval},
  year = {2009},
  pages = {758--759},
  doi = {10.1145/1571941.1572114}
}

@article{robinson2021hardnegative,
  title = {{Contrastive Learning with Hard Negative Samples}},
  author = {Joshua Robinson and Ching-Yao Chuang and Suvrit Sra and Stefanie Jegelka},
  journal = {arXiv preprint arXiv:2010.04592},
  year = {2020},
  doi = {10.48550/arXiv.2010.04592}
}

@inproceedings{chuang2020debiased,
  title = {{Debiased Contrastive Learning}},
  author = {Chuang, Ching-Yao and Robinson, Joshua and Lin, Yen-Chen and Torralba, Antonio and Jegelka, Stefanie},
  booktitle = {Advances in Neural Information Processing Systems},
  year = {2020},
  volume = {33},
  pages = {8765--8775},
  note = {\href{https://proceedings.neurips.cc/paper/2020/hash/63c3ddcc7b23daa1e42dc41f9a44a873-Abstract.html}{Open-access proceedings}}
}

@article{gulrajani2021domainbed,
  title = {{In Search of Lost Domain Generalization}},
  author = {Ishaan Gulrajani and David Lopez-Paz},
  journal = {arXiv preprint arXiv:2007.01434},
  year = {2020},
  doi = {10.48550/arXiv.2007.01434}
}

@article{hu2022lora,
  title = {{LoRA: Low-Rank Adaptation of Large Language Models}},
  author = {Edward J. Hu and Yelong Shen and Phillip Wallis and Zeyuan Allen-Zhu and Yuanzhi Li and Shean Wang and Lu Wang and Weizhu Chen},
  journal = {arXiv preprint arXiv:2106.09685},
  year = {2021},
  doi = {10.48550/arXiv.2106.09685}
}

@misc{aicity2026results,
  title = {{2026 Challenge Winners}},
  author = {{AI City Challenge Organizers}},
  year = {2026},
  howpublished = {\url{https://www.aicitychallenge.org/2026-challenge-winners/}},
  note = {Accessed October 22, 2026}
}
\end{document}